\documentclass[12pt]{article}

\usepackage{PRIMEarxiv}
\usepackage{multirow}
\usepackage[utf8]{inputenc} 
\usepackage[T1]{fontenc}    
\usepackage{hyperref}       
\usepackage{url}            
\usepackage[noadjust]{cite} 
\usepackage{booktabs}       
\usepackage{tabularx}       
\usepackage{array}
\usepackage{newunicodechar}
\newunicodechar{？}{?}
\usepackage{amsfonts}       
\usepackage{amsmath}        
\usepackage{nicefrac}       
\usepackage{microtype}      
\usepackage{lipsum}
\usepackage{fancyhdr}       
\usepackage{graphicx}       
\usepackage{xcolor}         
\usepackage{subcaption}
\usepackage{float}

\makeatletter
\renewcommand{\section}{%
  \@startsection{section}{1}{\z@}%
                {-1.8ex \@plus -0.5ex \@minus -0.2ex}%
                { 0.9ex \@plus  0.2ex \@minus  0.1ex}%
                {\large\bf\raggedright}%
}
\renewcommand{\subsection}{%
  \@startsection{subsection}{2}{\z@}%
                {-1.6ex \@plus -0.4ex \@minus -0.2ex}%
                { 0.55ex \@plus  0.15ex \@minus  0.1ex}%
                {\normalsize\bf\raggedright}%
}
\renewcommand{\subsubsection}{%
  \@startsection{subsubsection}{3}{\z@}%
                {-1.3ex \@plus -0.4ex \@minus -0.2ex}%
                { 0.35ex \@plus  0.15ex \@minus  0.1ex}%
                {\normalsize\bf\raggedright}%
}
\makeatother

\graphicspath{{media/}}     
\graphicspath{{figures/}}
\title{CANN~Bench: Benchmarking Agent Generated Kernels against Real NPU and Algorithmic Limits
}

\newcommand{\symfootnote}[2]{%
  \begingroup
  \renewcommand{\thefootnote}{#1}%
  \footnotetext{#2}%
  \addtocounter{footnote}{-1}%
  \endgroup
}

\author{%
  \parbox{0.95\textwidth}{\centering
    Xue-Jian Gao\textsuperscript{*}, Deng Pan\textsuperscript{*}, Yueming Su\textsuperscript{*},
    Jiasheng Li, Bin Du, Fengming Zhu, Chengdi Ma, Junyi Fan,
    Qichen Liao, Chengqiu Hu, Xinxian Chen, Lingchao Zheng, Jun Li, Jiwei Yang,
    Yuwei Fan\textsuperscript{\#}\\[0.4ex]
    \normalsize\normalfont Huawei
  }%
}

\begin{document}
\maketitle
\symfootnote{*}{~These authors contributed equally: Xue-Jian Gao, Deng Pan, Yueming Su}
\symfootnote{\#}{~To whom the correspondence should be addressed:  \texttt{fanyuwei2@huawei.com}}
\setcounter{footnote}{0}

\begin{abstract}
AI agents are now capable of writing, compiling, and iteratively optimizing low-level operator kernels on different hardware platforms. Existing benchmarks, however, focus almost exclusively on CUDA and Triton, leaving hardware ecosystems with less-exposed programming models without a common evaluation baseline. We present \textbf{CANN~Bench}, an open benchmark for AI-generated operator code on Huawei's Ascend NPU. The current release covers \textbf{53 operators} and \textbf{1060 test cases} organized into four difficulty tiers---from simple elementwise primitives to MoE dispatch and FlashAttention kernels---spanning FP16, BF16, FP32, and INT8 precision formats. Evaluation adopts a \textbf{three-dimensional weighted composite score} that treats compilation, functional correctness, and performance as independent axes, providing a principled reward signal for kernel-generation agents. Performance is graded against an out-of-the-box PyTorch-on-Ascend baseline and an analytical per-case Hardware-Anchored Performance (HAP) limit on real NPU hardware, ensuring scores reflect genuine optimization headroom rather than measurement artifacts. The evaluation harness is designed to resist reward hacking from the ground up. CANN~Bench is versioned within the official CANN repository and is designed for long-term community co-construction, providing the Ascend ecosystem with a quantitative, reproducible, and sustainably maintained yardstick for AI operator-authoring capability.
\end{abstract}



\section{Introduction}
\label{sec:intro}


\subsection{Why an Ascend-Oriented Benchmark Now}
\label{sec:why-bench}

Code-generation agents now write, compile, and profile operator kernels end to end~\cite{cudaagent2026,cutegen2026,autokernel2026,gpukernelscientist2025,geak2025}, and the benchmarks they target supply the execution-feedback signal that drives each next round of optimization~\cite{kevin2025,drkernel2026,evoengineer2025,lange2025robust}. On Huawei's Ascend NPU, kernel generation runs through CANN---Huawei's heterogeneous computing architecture that bridges upper-layer AI frameworks and the underlying Ascend processors. Writing a kernel under CANN requires explicit control of the memory hierarchy, tile shapes, and Vector--Cube pipeline scheduling, a programming surface analogous to CUDA on the NVIDIA GPU architecture. Recent benchmarks have begun to include Ascend targets~\cite{multikernelbench2025,cao2026ascendkernelgen,ascendcraft2026,ascendoptimizer2026,evokernel2026,kernelcat2025}, but they either treat Ascend as one platform among many or are coupled to a specific agent-training recipe. The Ascend ecosystem therefore still lacks a shared evaluation scale for AI-generated CANN kernels---one independent of any single training recipe and maintained as a cross-project reference.

The catalog of operators that must be supported continues to expand. Each new model family introduces new attention variants and MoE routing patterns that have to be re-tiled and re-scheduled for Ascend under CANN. Precision compounds the growth: modern pipelines mix FP16 and BF16 with low-precision formats such as FP8 and MXFP4~\cite{rouhani2023microscaling,tseng2025mxfp4}, each with distinct granularity and performance trade-offs. Workload churn and precision growth together push the operator catalog beyond what hand-authored kernels can track. An Ascend-oriented benchmark therefore needs to evolve with the hardware and software stack rather than remain fixed at release. We define six design criteria for such a benchmark below.

\subsection{Six Criteria for an Ascend-Oriented Benchmark}
\label{sec:six-criteria}

We organize the six criteria into two layers. The first four concern the task set; the remaining two concern benchmark delivery and maintenance.

\noindent\textbf{C1. Difficulty span.} The task set must span the full difficulty range, from element-wise and reduction primitives, through compute-bound cores such as convolution and matrix-multiplication, up to fusion-level operators such as MoE dispatch and Flash Attention. Without this range, scores cluster at the trivial or fusion-only extremes and fail to separate capability tiers.

\noindent\textbf{C2. Workload provenance.} Tasks come from real Ascend production workloads---LLM training, quantized inference, and vision/multimodality pipelines---rather than from synthetic cases whose failure modes do not transfer.

\noindent\textbf{C3. Precision and quantization coverage.} FP16 and BF16 are the baseline. Quantized integer paths (INT8 with per-tensor and per-channel granularities) are equally important in production under CANN, and lower-precision floating-point formats such as FP8, HiF8, MXFP8, FP4, and MXFP4~\cite{rouhani2023microscaling,tseng2025mxfp4} are on the near-term roadmap (\S\ref{sec:status}).

\noindent\textbf{C4. Evaluation on three axes.} Compilation, functional correctness, and performance on Ascend NPUs all contribute to the final weighted score; failing any one of them strictly limits the credit a submission can receive on the remaining axes.

\noindent\textbf{C5. Self-contained reproducibility kit.} Each operator ships with a specification, a golden implementation, public test cases, and a performance baseline, so that third parties can reproduce the evaluation.

\noindent\textbf{C6. Anti-reward-hacking and sustained evolution.} The harness must detect reward-hacking attempts such as bypassing accuracy assertions, memorizing outputs, or hard-coding shapes. The task set and evaluation protocol must also evolve with Ascend hardware, CANN releases, and model families rather than remain fixed at release.

\subsection{Existing Landscape Against These Criteria}
\label{sec:landscape}

Measured against the criteria in \S\ref{sec:six-criteria}, existing benchmarks cover only subsets of the design space. We group the closest prior work by target hardware and summarize their positions in Table~\ref{tab:related-compare}.

The CUDA and Triton ecosystem hosts the largest body of kernel-generation benchmarks. KernelBench~\cite{ouyang2025kernelbench} decomposes evaluation into compilation, correctness, and speedup across 250 PyTorch-to-CUDA tasks. TritonBench~\cite{li2025tritonbench} applies a similar three-axis setup to 184 real-world Triton operators, where DSL-concept hallucination is a dominant failure mode. SOL-ExecBench~\cite{lin2026solexecbench} anchors performance to hardware Speed-of-Light bounds on NVIDIA Blackwell and adds a sandboxed harness for timing. All three target NVIDIA platforms. Their task sets do not transfer directly to Ascend: CANN exposes different tiling semantics and a different multi-level buffer structure, so a mechanical port fails C2 (workload provenance).

The same non-transfer appears on other NPU stacks. NPUEval~\cite{kalade2025npueval} evaluates frontier LLMs on 102 AMD-NPU kernel tasks and reports an average vectorization score of around 10\%. KernelCraft~\cite{kernelcraft2026} studies close-to-metal kernel generation across the PLENA, AMD NPU, and Coral NPU ISAs on 33 tasks. Neither targets CANN nor Ascend.

Two benchmarks target the Ascend ecosystem more directly, but neither fully satisfies our six criteria. MultiKernelBench~\cite{multikernelbench2025} covers 285 tasks across 14 kernel categories and three hardware platforms---NVIDIA GPUs, Huawei Ascend NPUs, and Google TPUs---using author-provided references and a category-aware one-shot prompting scheme. Breadth costs per-platform depth: Ascend tasks are one slice of the total, so per-operator specification depth is limited (partial C5) and Ascend precision coverage is narrow (partial C3). NPUKernelBench, released alongside the AscendKernelGen fine-tuning recipe~\cite{cao2026ascendkernelgen}, covers Ascend tasks at three difficulty levels; AscendKernelGen reports a Level-2 compile rate rising from 0\% to 95.5\% Pass@10 with chain-of-thought data plus reinforcement learning from on-device execution feedback. Two limitations weaken its role as a shared reference. First, on the static-shape track that constitutes a substantial portion of NPUKernelBench, all test cases for a given task correspond to a single fixed input configuration~\cite{cao2026ascendkernelgen}, so a kernel specialized to one canonical shape already passes per-task evaluation. Second, the benchmark is released and evolved alongside the AscendKernelGen recipe rather than on an independent maintenance track, so its task set tracks the cadence of one training pipeline rather than the broader Ascend ecosystem (partial C6).

Several kernel-generation methods now target CANN directly. AscendCraft~\cite{ascendcraft2026} applies DSL-guided transcompilation through a constraint-directed intermediate representation. AscendOptimizer~\cite{ascendoptimizer2026} runs an episodic evolutionary agent with optimization-rewind over 127 operators from the \texttt{cann-ops} repository and reports a 1.19$\times$ geometric-mean speedup. EvoKernel~\cite{evokernel2026} addresses cold-start and continual refinement through value-driven memory; on a self-constructed NPU variant of KernelBench it lifts frontier-model correctness from 11\% to 83\%. KernelCAT~\cite{kernelcat2025} reports deployment-side gains on FlashAttention and DeepSeek-OCR-2 on Ascend 910B2. Each of these efforts uses either a cross-platform benchmark, a custom task set, or an author-chosen operator collection, so they still lack a common evaluation scale.

C6 (anti-reward-hacking and sustained evolution) has a concrete precedent. KernelBench has documented silent-dispatch exploits in practice: submissions that call high-level operators such as \texttt{cuBLAS} in place of the generated CUDA kernel, or no-op kernels whose output memory is reused from the reference tensor~\cite{kernelbench_v01_blog,rewardhackdefense2025,kernelbenchissue74}. Table~\ref{tab:related-compare} positions the above benchmarks along five axes: golden-reference provenance, scoring format, generalization discipline, anti-reward-hacking stance, and maintenance mode.
Among publicly described Ascend NPU benchmarks, we are not aware of another that combines co-location in the official CANN repository, maintenance alongside vendor operator contracts, and a dedicated leaderboard protocol; CANN~Bench is designed to cover all three.

\begin{table}[t]
\centering
\footnotesize
\renewcommand{\arraystretch}{1.12}
\setlength{\tabcolsep}{3pt}
\begin{tabularx}{\textwidth}{@{}>{\raggedright\arraybackslash}p{0.21\textwidth} >{\raggedright\arraybackslash}X >{\raggedright\arraybackslash}X >{\raggedright\arraybackslash}X >{\raggedright\arraybackslash}X >{\raggedright\arraybackslash}p{0.125\textwidth}@{}}
\toprule
\textbf{Benchmark} & \textbf{Golden Provenance} & \textbf{Scoring / Performance Metric} & \textbf{Test-case Generalization} & \textbf{Anti-Reward-Hacking} & \textbf{Maintenance} \\
\midrule
MultiKernelBench~\cite{multikernelbench2025} (Huawei NPU tasks) & Author-provided reference & Compile / Correctness / Speedup & --- & --- & Academic \\
\addlinespace[3pt]
NPUKernelBench~\cite{cao2026ascendkernelgen} & Author-provided reference & Compile / Correctness / Performance, reported as Pass@$k$ & --- & --- & Academic (paired w/ AscendKernelGen) \\
\addlinespace[3pt]
AscendOptimizer bench~\cite{ascendoptimizer2026} & \texttt{cann-ops} operators & Speedup only (optimization stage) & --- & --- & Academic \\
\addlinespace[3pt]
\textbf{CANN~Bench (ours)} & \textbf{Vendor-authoritative (official CANN repo)} & \textbf{Three-dimensional composite score (see the Evaluation section)} & \textbf{20 public + 80 hidden cases per op} & \textbf{Submission rules + Python/ATen guards + timing integrity + NPU-kernel scoring} & \textbf{CANN repo, versioned; online leaderboard (forthcoming)} \\
\bottomrule
\end{tabularx}
\caption{Ascend ecosystem kernel-generation benchmarks: positioning of CANN~Bench against relevant benchmarks targeting Ascend NPUs. Benchmarks targeting other ISAs (KernelBench, TritonBench, SOL-ExecBench, KernelCraft, NPUEval) are omitted from direct comparison. Columns report provenance of the golden reference, performance-score composition, test-case generalization discipline, anti-reward-hacking stance, and maintenance mode; ``---'' denotes an attribute that is not documented for a given benchmark.}
\label{tab:related-compare}
\end{table}

\subsection{CANN~Bench: Dataset, Scoring, and Online Leaderboard}
\label{sec:cannbench-overview}
CANN~Bench is an operator-generation benchmark maintained in the official CANN repository. Its specifications, golden implementations, test cases, and performance baselines sit alongside Huawei's vendor-maintained operator contracts (C2, C5). The initial release targets Ascend 910B2 with a kernel-DSL-agnostic specification format designed to extend to Triton, PyPTO, TileLang, and other CANN-compatible paradigms in future releases.

\paragraph{Dataset.} CANN~Bench currently contains 53 operators and 1060 test cases, distributed 8/16/21/8 across four difficulty levels (L1--L4) that track Ascend hardware-unit usage, from Vector-only kernels at L1 to fused Matrix--Vector pipelines at L4. \S\ref{sec:benchmark} lists representative operators, the full per-level breakdown, and the four per-operator reproducibility artifacts (\texttt{desc.md}, \texttt{proto.yaml}, \texttt{cases.csv}, \texttt{golden.py}).

\paragraph{Public and hidden case split.} Each operator's cases split into a 20-case public set for local iteration and an 80-case hidden set that the harness retains for leaderboard evaluation. The leaderboard aggregates both splits, so the hidden set guards against fit to disclosed inputs alone.

\paragraph{Three-dimensional weighted scoring.} Compilation acts at the operator level, while functional correctness and performance are scored case by case; the three signals are combined into a per-operator weighted composite (\S\ref{sec:composite-score}),
$$
  \mathrm{EachOperatorScore}
  = [w_c \cdot \delta_\mathrm{pass} + \sum_{i \in \mathrm{cases}}
  \frac{                            
  \delta_{\mathrm{accuracy},i} (w_f+ w_p\cdot   
  \mathrm{score}_i)}{\mathrm{num\_of\_cases}} ]\cdot 100,          
$$
where $\delta_{\mathrm{pass}} \in \{0,1\}$ is the operator-level compilation flag (a kernel that fails to compile zeros out every term), $\delta_{\mathrm{accuracy}, i} \in \{0,1\}$ is the per-case functional indicator, $\mathrm{score}_i$ is the per-case performance sub-score, and $w_c, w_f, w_p$ are linear weights (default $w_c = 0.2$, $w_f = 0.3$, $w_p = 0.5$). The performance sub-score is the HAP score: it grades the candidate kernel's measured runtime $T_{\text{cand},i}$ between a PyTorch-on-Ascend baseline $T_{\text{baseline},i}$ and an analytically derived per-case Hardware-Anchored Performance (HAP) limit $T_{\text{HW},i}$ (\S\ref{sec:hwlimit}, \S\ref{sec:scoring}).

\paragraph{Precision scope.} The current release covers FP16, BF16, FP32, and quantized integer paths (notably INT8 for weight and activation quantization). FP8, HiF8, MXFP8, FP4, and MXFP4 remain planned extensions discussed in \S\ref{sec:status}.

\paragraph{Online leaderboard.} CANN~Bench is delivered with an online leaderboard\footnote{The portal is being prepared and will be released in a future update.}.

\subsection{Engineering Substrate and Anti-Reward-Hacking}
\label{sec:substrate}
Reliable scoring depends on the harness. CANN~Bench runs compilation, accuracy verification, and performance sampling in a single pipeline. To control measurement noise on the NPU path, it issues a heavy MatMul$+$ReduceMax warmup that drives the NPU to its sustained peak DVFS state, evicts the L2 cache between timed runs to neutralize cross-step cache reuse, and collects kernel-level timings through the \texttt{torch\_npu.profiler} surface, which exposes the same per-kernel metrics as the Ascend \texttt{msprof} toolchain~\cite{msprof_ascend}. To isolate cross-case state, it executes each operator in a fresh subprocess. Performance baselines are versioned alongside each task in \texttt{cases.yaml} rather than frozen at release: the values shipped today reflect the CANN release used to record them, and a refresh utility that regenerates \texttt{baseline\_perf\_us} as hardware and CANN versions evolve is slated for an upcoming release, so speedups can be re-anchored to the current attainable reference rather than a stale snapshot.

Our reward-hacking defense follows a layered model rather than one monolithic check. The submission contract rules out high-level compute delegation, same-op routing to preinstalled CANN kernels, CPU fallback, output caching, fake or lazy tensors, and timing-API tampering. At runtime, the evaluator enforces this contract around framework calls, operator dispatch, device residency, and measurement integrity, while the scorer grants no performance credit when a correctness-passing run cannot be attributed to candidate NPU-kernel execution. Numerical validation remains dtype-aware, and leaderboard scores aggregate over both public and hidden splits, so a submission cannot be tuned to disclosed inputs alone.

Two surfaces remain deliberately conservative. Multi-stream submissions are restricted rather than fully ranked because stable all-stream visibility across \texttt{torch\_npu} versions is still fragile, and opaque pre-compiled binary loading remains a manual-review item. CANN~Bench also supports evaluator-machine hardening against same-op library routing on dedicated evaluation hosts. The mitigation principles and remaining limitations are detailed in \S\ref{sec:reward-hacking}.

\subsection{Status, Roadmap, and Open-Source Plan}
\label{sec:status}
As of this release, CANN~Bench provides specifications, golden implementations, public test cases, and performance baselines for 53 operators across L1--L4. The harness and baseline-collection pipeline have completed end-to-end runs on NPU hardware. The public leaderboard is being prepared and will be released in a future update.

Three extensions are planned. The leaderboard portal will index and date every submission so that results stay directly citable. Kernel-DSL coverage will extend the initial Ascend-optimized support to Triton, PyPTO, TileLang, and other CANN-compatible paradigms under the same specification and scoring protocol. Precision coverage will expand the current FP16 / BF16 / FP32 / INT8 baseline to FP8, MXFP8, FP4, MXFP4, HiF8, and related low-precision formats as operators and baselines are validated.

The operator set, golden implementations, harness, and documentation are open-source. The project accepts contributions of new operators, test cases, baseline updates, and sub-leaderboards, so the benchmark tracks Ascend hardware and workload evolution over time.

\paragraph{Availability and license.} CANN~Bench is hosted in the official CANN repository at \url{https://gitcode.com/cann/cann-bench} and released under the \emph{CANN Open Software License Agreement Version 2.0}~\cite{cann_license_v2}. The harness targets Ascend 910B2 and is validated against CANN~$\geq 9.0.0$, the version we recommend for reproducing the reported baselines; older toolkit releases are not guaranteed to match the published \texttt{baseline\_perf\_us} values, which are versioned alongside the harness in each operator's \texttt{cases.yaml} and will be refreshable via a baseline-update utility scheduled for an upcoming release.

\section{Benchmark and Dataset}
\label{sec:benchmark}

\subsection{Design Principles}

\textbf{CANN~Bench} is organized around three principles for Ascend C kernel evaluation.

\begin{enumerate}

\item \textbf{Vendor-authoritative provenance.}
Every workload in CANN~Bench draws on official artifacts from the CANN repository, including standardized operator definitions, reference implementations, and standard test cases. CANN~Bench is distinct from the vendor-maintained operator base \texttt{opbase}\footnote{\url{https://gitcode.com/cann/opbase}.}: \texttt{opbase} is the production operator collection that ships with CANN, while CANN~Bench is an evaluation benchmark for AI-generated kernels whose task set partially overlaps with \texttt{opbase} but is not identical. Even on shared tasks, CANN~Bench specifications are not transcribed verbatim: we relax a subset of the original functional requirements---for example, narrowing the supported dtype and shape envelope or trimming peripheral code paths---so the evaluation focuses on the core kernel logic rather than the surrounding production machinery. Beyond these canonical operators, the benchmark also includes emerging custom operators absent from the official codebase, such as \texttt{Engram} from DeepSeek~\cite{cheng2026conditionalmemoryscalablelookup}. Combining canonical and newly proposed operators broadens the scope on which generation and optimization can be evaluated.

\item \textbf{Four levels with increasing difficulty.}
Operators are grouped into four hierarchical levels (L1--L4), from element-wise kernels at L1 through progressively composite, contraction-heavy, and fused production kernels at L4.
The levels also map to a hardware-utilization progression: L1 and L2 use only vector operations, L3 introduces basic combinations of vector and cube operations, and L4 requires tightly coordinated vector--cube pipelines within a single fused kernel. Table~\ref{tab:benchmark-overview} summarizes the four levels.
Our levels reflect how each operator maps onto Ascend's execution model, not shallow features such as code length or operator arity that prior kernel-generation benchmarks have used as difficulty proxies. The level at which an agent first stalls thus localizes its capability gap: a system that clears L1 and L2 cleanly but degrades sharply at L3 reveals a specific weakness on cube-side reasoning rather than producing only a lower aggregate score. The resulting per-level profile delivers the capability-tier separation that C1 calls for, not a single-axis difficulty score.

\item \textbf{Case diversity rather than single-shape scoring.}
Each operator carries a family of concrete cases that vary tensor shape, dtype, and attribute settings.
The design avoids rewarding kernels that overfit a single canonical configuration and instead tests whether generated code generalizes across the combinations that matter in real Ascend C development.
Across the current split, the cases cover float16 / bfloat16 / float32 plus integer paths (int8, uint8, int32, int64) and mixed-dtype configurations.
\end{enumerate}

\subsection{Benchmark Structure and Dataset Overview}

\begin{figure}[t]
  \centering
  \begin{subfigure}[t]{0.46\textwidth}
    \includegraphics[width=\textwidth]{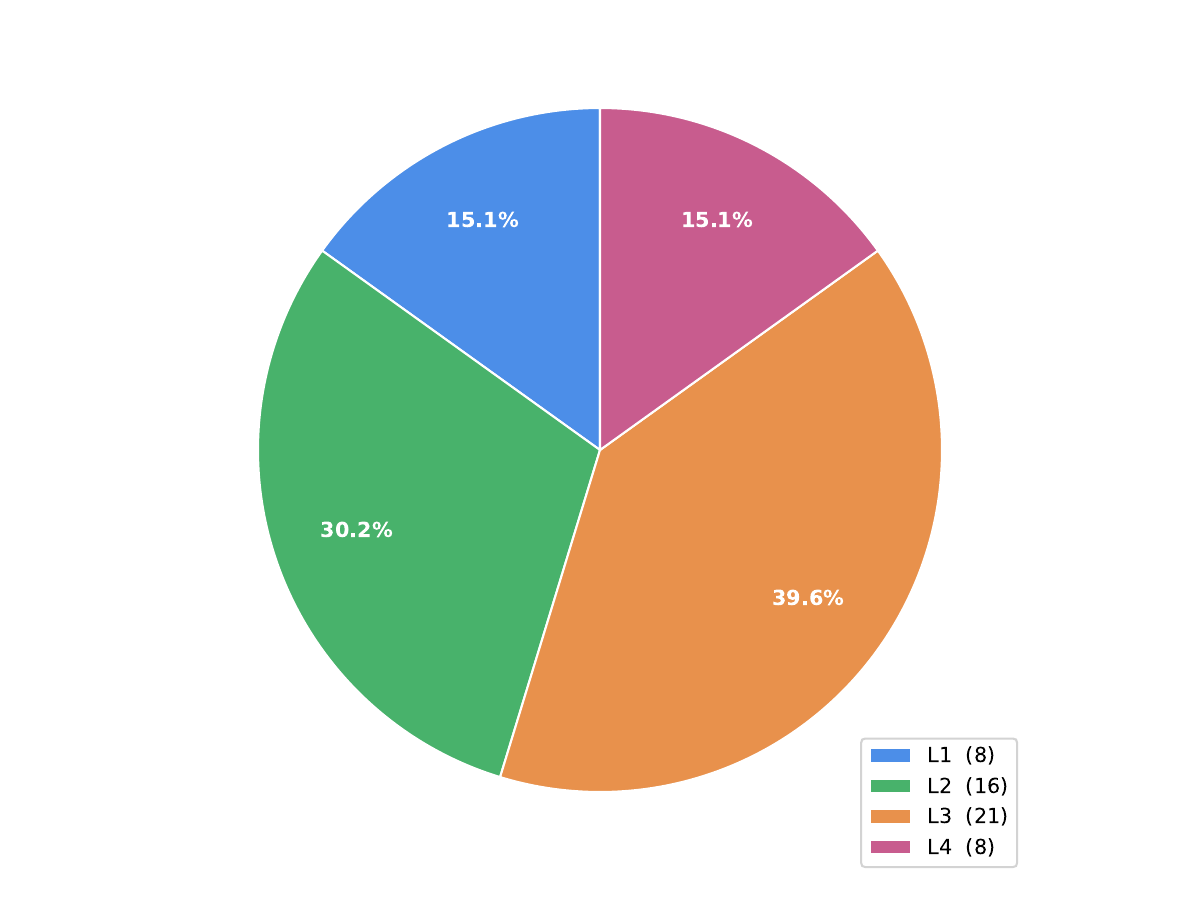}
    \caption{Operator count per difficulty level.}
    \label{fig:dataset-a}
  \end{subfigure}
  \hfill
  \begin{subfigure}[t]{0.46\textwidth}
    \includegraphics[width=\textwidth]{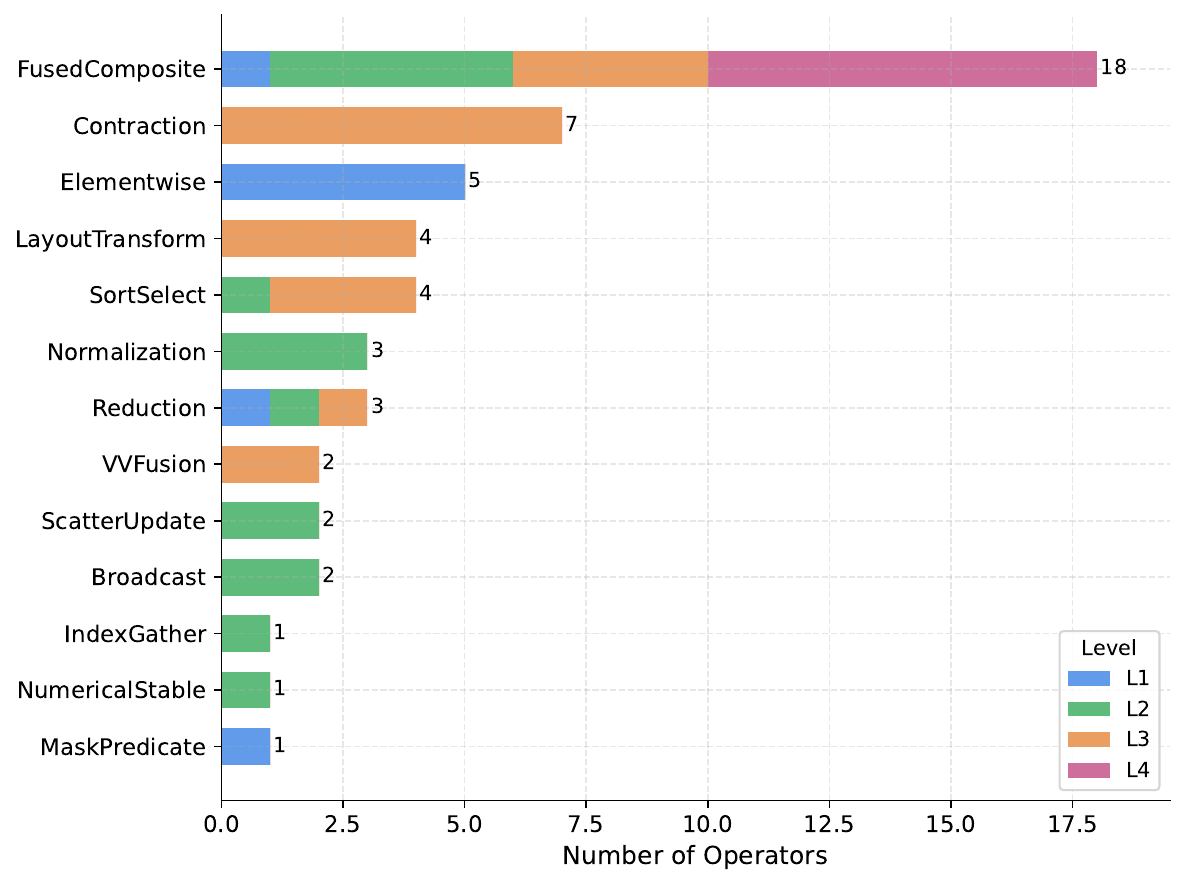}
    \caption{Operation category distribution by level.}
    \label{fig:dataset-b}
  \end{subfigure}

  \vspace{6pt}

  \begin{subfigure}[t]{0.46\textwidth}
    \includegraphics[width=\textwidth]{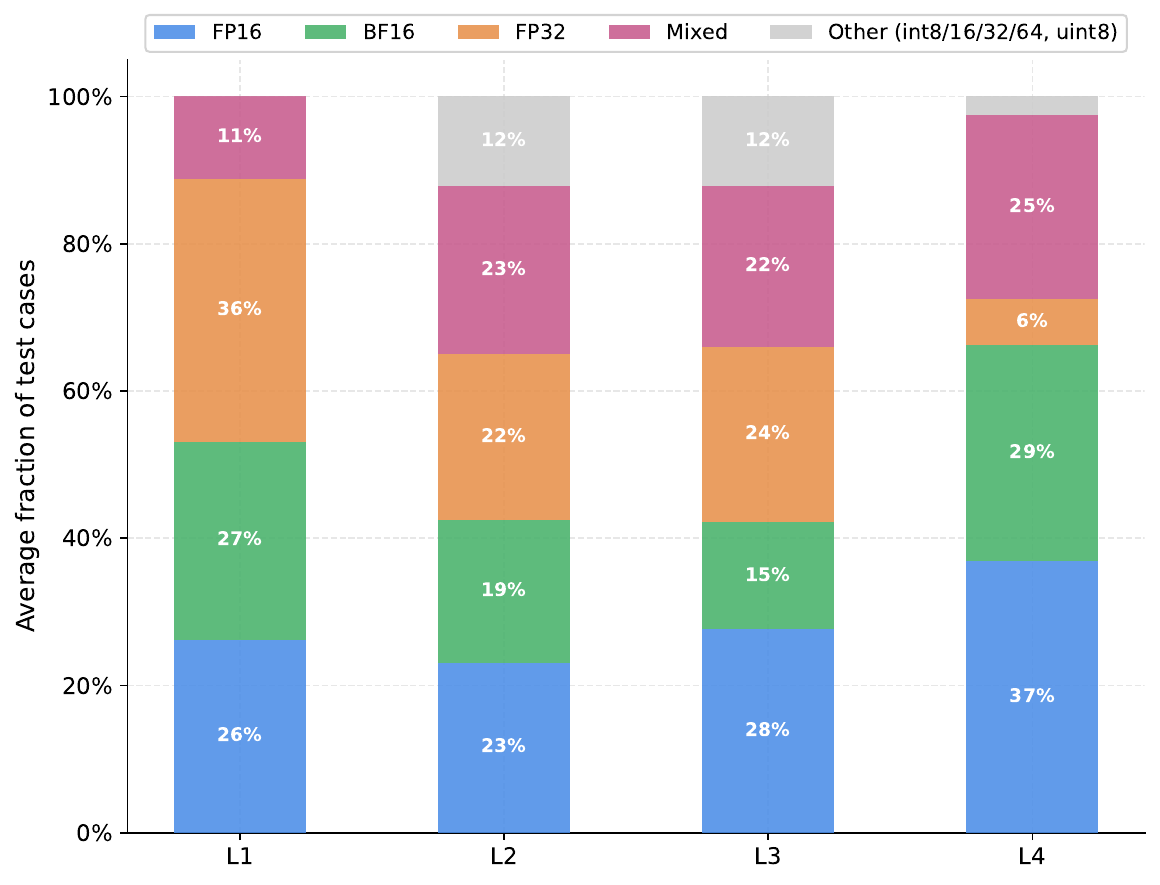}
    \caption{Primary compute dtype distribution per level.}
    \label{fig:dataset-d}
  \end{subfigure}
  \hfill
  \begin{subfigure}[t]{0.46\textwidth}
    \includegraphics[width=\textwidth]{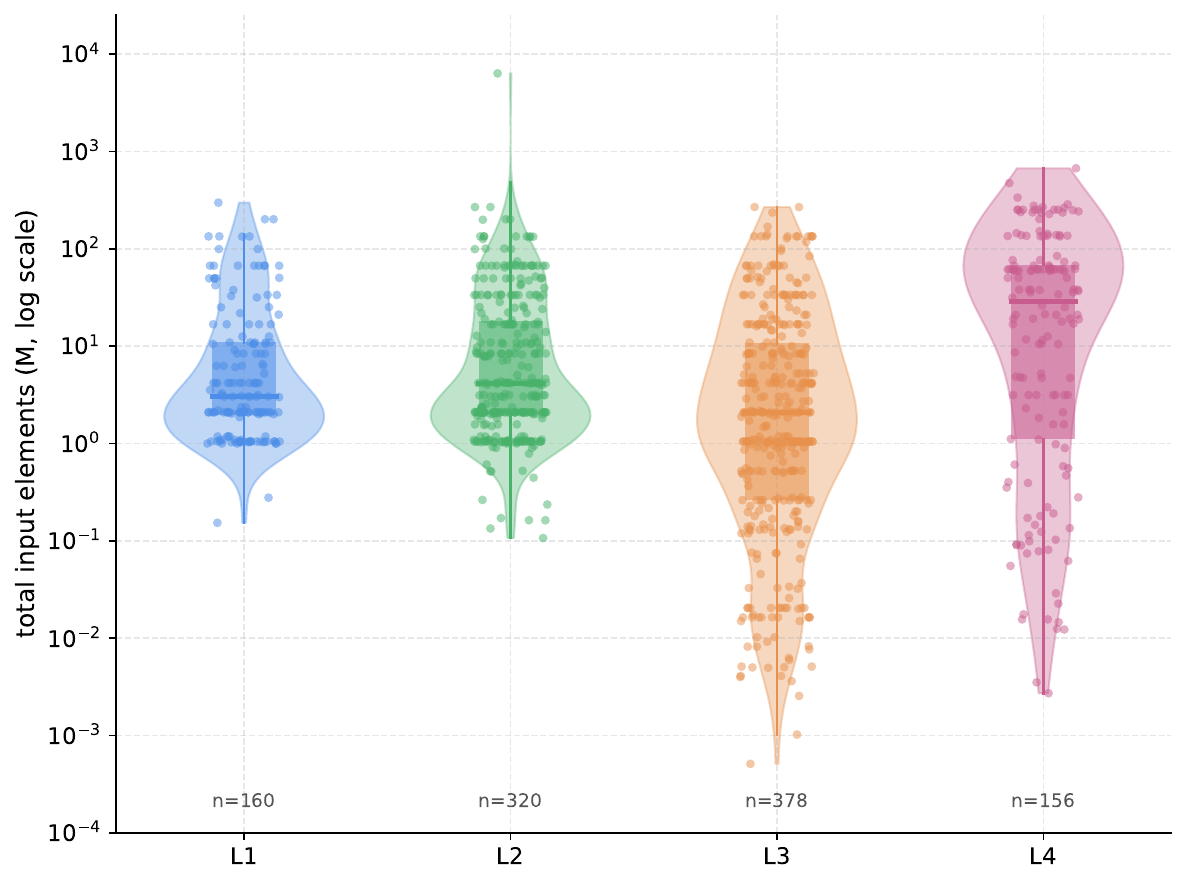}
    \caption{Total number of elements of the test cases per level.}
    \label{fig:dataset-f}
  \end{subfigure}

  \caption{Dataset overview of our CANN~Bench.
  (\subref{fig:dataset-a}) Operator count per difficulty level.
  (\subref{fig:dataset-b}) Operation category distribution, stacked by level contribution.
  (\subref{fig:dataset-d}) Distribution of the primary compute data type per level.
  (\subref{fig:dataset-f}) Distribution of the provided baseline execution times (log-scale).}
  \label{fig:dataset}
\end{figure}

Table~\ref{tab:benchmark-overview} summarizes the current CANN~Bench release: \textbf{53 operators}, each with about 20 test cases.
The benchmark is versioned alongside the CANN repository and grows as additional operators are promoted from the experimental track into the curated set. Figure~\ref{fig:dataset} complements Table~\ref{tab:benchmark-overview} with four views of the 1060 test cases.

\begin{table}[t]
\centering
\captionsetup{width=0.85\linewidth}  
\renewcommand{\arraystretch}{1.12}
\setlength{\tabcolsep}{4pt}
\small
\begin{tabular}{@{}cc>{\raggedright\arraybackslash}p{4.7cm}>{\centering\arraybackslash}p{1.45cm}>{\raggedright\arraybackslash}p{5.2cm}@{}}
\toprule
Level & Count & Description & Type & Representative Operators \\
\midrule
L1 & 8  & Single-primitive kernels involving only Vector Core operations & Vector & \texttt{gelu}, \texttt{swi\_glu}, \texttt{sigmoid}, \texttt{mish} \\
L2 & 16 & Fused composite kernels involving complex Vector Core operations without matrix contractions & Vector & \texttt{rms\_norm}, \texttt{cross\_entropy\_loss}, \texttt{softmax}, \texttt{apply\_rotary\_pos\_emb} \\
L3 & 21 & Kernels combining matrix contractions with Vector Core operations & Vector \& Cube & \texttt{conv2\_d}, \texttt{grouped\_matmul}, \texttt{top\_k}, \texttt{dequant\_swiglu\_quant} \\
L4 & 8  & Complex fused kernels requiring coordinated Matrix and Vector Core pipelines within a single kernel & Vector \& Cube & \texttt{sparse\_flash\_attention}, \texttt{mla\_prolog}, \texttt{gru}, \texttt{lstm} \\
\midrule
\textbf{Total} & \textbf{53} & & & \\
\bottomrule
\end{tabular}
\caption{A summary of CANN~Bench. The four levels track a progression in implementation difficulty, from local pointwise kernels to production-grade fused pipelines.}
\label{tab:benchmark-overview}
\end{table}

\paragraph{Level distribution (Figure~\ref{fig:dataset-a}).}
Operators are distributed across four hierarchical levels as mentioned above. The third level contains the most operators (21), spanning contraction, sorting, layout conversion, and fused quantization kernels under a heterogeneous mix of vector and cube workloads. The second level follows (16), highlighting the widespread use of multi-input normalization, indexing, and fused utility kernels in practical Ascend workloads. The first and fourth levels are tied for the fewest (8 each): L1 concentrates on single-primitive vector kernels, while L4 gathers the most engineering-heavy fused pipelines under the current classification.

\paragraph{Operation category distribution (Figure~\ref{fig:dataset-b}).}
The category labels span a wide range of operator types.
\emph{FusedComposite} is the largest category, followed by \emph{Contraction}, \emph{Elementwise}, and \emph{SortSelect}.
Composition differs across levels: L1 centers on elementwise kernels, L2 on fused composite and normalization operators, L3 on contraction and reordering-heavy kernels, and L4 consists entirely of fused composite operators.
The distribution shows that CANN~Bench extends beyond pure GEMM evaluation to indexing, reduction, layout manipulation, and control-intensive operators---common failure points for AI-assisted Ascend C code generation.

\paragraph{Data-type distribution (Figure~\ref{fig:dataset-d}).}
The figure reports the primary compute dtype of each operator. L1 and L2 are dominated by float16, bfloat16 and float32 paths, which matches the typical workload of activation, normalization and utility kernels. L3 and L4 contain a larger share of mixed-dtype and integer cases, as several operators in those tiers fuse floating-point activations with quantized weights, integer metadata, and multi-input specifications.


\subsection{Test-Case Design Methodology}
\label{sec:case-design}

Each operator in CANN~Bench carries a structured case set rather than a hand-picked collection of inputs. The construction follows four explicit coverage axes and ends with per-operator uniqueness constraints, so each case contributes evidence the others do not.

\paragraph{Tensor-size coverage(Figure~\ref{fig:dataset-f}).}
Each case is sized by its total \emph{input element count}---the sum of $\prod_i d_i$ over all of its input tensors---which spans roughly $10^6$ to $3{\times}10^8$. The benchmark covers this range across three regimes: small cases (a few million elements) sit inside the on-chip buffers and surface tiling and launch-overhead issues, large cases (around $10^8$ and above) force off-chip memory traffic and expose bandwidth-bound behavior, and a medium band (tens of millions) covers the bulk of production shapes. A kernel that overfits to a single regime cannot collect the per-case credit at the other two.

\paragraph{Axis alignment.}
We label a shape \emph{aligned} when the byte length of its innermost dimension is a multiple of \textbf{512\,B}, and \emph{non-aligned} otherwise. The threshold is grounded in the Ascend MTE (Memory Transfer Engine) DMA path: a 512\,B-aligned innermost stride lets MTE issue full-burst transfers between global memory and the on-chip Unified Buffer at peak bandwidth, whereas a non-aligned trailing dimension forces MTE to break the transfer into a head/tail pair of slower unaligned bursts plus masked tail handling, materially degrading effective bandwidth even when functional correctness is preserved. Concretely, a 512\,B boundary corresponds to $256$ elements for FP16/BF16, $128$ elements for FP32, and $512$ elements for INT8; common ML-friendly shapes such as $1024$, $2048$, or $4096$ in the innermost axis fall on the boundary, while a prime or odd innermost dimension generally does not. Within each operator's case set we require \emph{at least 50\% non-aligned cases}, where the ratio is computed over the operator's 20 public cases and a case counts as non-aligned whenever the byte length of any input tensor's innermost dimension is not a multiple of 512\,B. Inside the non-aligned subset we further bias toward primes (e.g.\ $1009$, $1021$, $1537$, $4097$, $1{,}000{,}003$) so that residual sizes do not collapse onto convenient divisors that an aggressive auto-tiler might still find friendly. This is the operational definition behind the alignment-stress requirement referenced in \S\ref{sec:substrate}.

\paragraph{Dimensionality.}
The case set spans 1D through 5D inputs rather than freezing each operator at a single rank. Mixing ranks stresses tiling along axes other than the canonical innermost one and surfaces broadcast and reshape paths that a 2D-only or 3D-only test would not.

\paragraph{Value-range sweep.}
Inputs are drawn from a fixed taxonomy (Table~\ref{tab:value-range-strategy}) that covers symmetric and asymmetric ranges of three magnitudes, dtype boundaries (e.g.\ $\pm 65504$ for fp16, $\pm 88$ for the fp32 exponent), the special values $\pm\infty$ and $\mathrm{NaN}$, and the all-zero degenerate case. For ranges containing $\pm\infty$, the data generator injects the corresponding extremes into the head and tail of the flattened tensor while filling the interior with regular random values, so that overflow handling, denormal flushing, and special-value propagation are observable side-by-side with normal arithmetic.

\begin{table}[t]
\centering
\small
\renewcommand{\arraystretch}{1.15}
\begin{tabular}{@{}>{\raggedright\arraybackslash}p{3.4cm}>{\raggedright\arraybackslash}p{6.0cm}>{\raggedright\arraybackslash}p{4.6cm}@{}}
\toprule
Category & Example range & Probes \\
\midrule
Symmetric small  & $[-1, 1]$, $[-0.1, 0.1]$ & Typical activation regime \\
Symmetric medium & $[-10, 10]$, $[-100, 100]$ & Wider numerical range \\
Symmetric large  & $[-1000, 1000]$ & Saturation, accumulator pressure \\
Asymmetric       & $[-1, 2]$, $[-5, 10]$ & Skewed distributions \\
Type boundary    & fp16 $[-65504, 65504]$, fp32 $[-3.4\mathrm{e}38, 3.4\mathrm{e}38])$ & Overflow / underflow \\
Special values   & $\pm\infty$, $\mathrm{NaN}$ & Numerical stability \\
Zero             & $[0, 0]$ & Degenerate input path \\
\bottomrule
\end{tabular}
\caption{Value-range categories that populate test-case inputs. Each operator's 20-case public split is required to draw from at least the symmetric-small, type-boundary, and special-value groups, so that no operator is evaluated only on benign data.}
\label{tab:value-range-strategy}
\end{table}

\paragraph{Uniqueness constraints.}
Within each operator's 20-case public split we require three properties to be unique across cases: total input element count, the JSON-serialized attribute combination, and the value range. The constraint blocks a common failure mode in hand-curated benchmarks where two cases differ only cosmetically and effectively count once. A self-check loop with at least three rounds of validation runs in the contributing pipeline before a new case set lands in the repository.

\paragraph{Public and hidden splits.}
The 20 public cases per operator ship in the open-source repository and support local iteration, inspection, and reproducibility. An additional 80 hidden cases per operator are \emph{not} open-sourced: they reside inside the leaderboard's server-side evaluation backend, where submissions are scored after upload. Hidden cases follow the same coverage spec and uniqueness constraints as the public split, so a kernel tuned only against the public set has no informational shortcut to the hidden one. Keeping the hidden split off-repo is deliberate---releasing it would collapse the split into a single 100-case public set and remove the protection against fitting to disclosed inputs that is its entire purpose. The submission backend is being prepared together with the online leaderboard and will be released in a future update.

\subsection{Problem Specification Format}

Each task ships with four artifacts; SOL-ExecBench~\cite{lin2026solexecbench} adopts a similar but smaller set.

\begin{enumerate}
    \item \textbf{Operator specification (\texttt{desc.md}).}
A structured description covering the mathematical definition, interface contract (input/output shapes, dtypes, and constraints), and numerical accuracy tolerance.

    \item \textbf{Operator prototype (\texttt{proto.yaml}).}
A machine-readable YAML file capturing the operator name, category, schema, attributes, shape support, and supported data types for inputs and outputs.

    \item \textbf{Test cases (\texttt{cases.csv}).}
A CSV table whose rows are concrete test cases, each specifying input shapes, data types, hyper-parameters, and the baseline NPU execution time in microseconds.

    \item \textbf{Golden implementation (\texttt{golden.py}).}
A Python reference implementation, typically in PyTorch, that produces the expected output for each test case.
\end{enumerate}

Together, the four difficulty levels, the case-design methodology, and the four-artifact specification format fix the dataset surface that any submission is evaluated against: which operators are in scope, which configurations are tested per operator, and what each task hands the candidate kernel as input. The next section turns from the dataset to the scoring side: how the harness converts a submission's behavior on these artifacts into compilation, functional, and performance signals, and how those signals are combined into a per-operator and benchmark-wide composite score.

\section{Evaluation}
\label{sec:evaluation}

CANN~Bench scores each test case on three axes: compilation, functional correctness, and performance. 
The first two are binary gates; performance is continuous, anchored to a hardware-derived bound rather than a moving framework baseline. 
Linear weights combine the three into a per-operator composite.

For each test case, three reference times are recorded:
$T_{\text{baseline}}$, the measured runtime of the operator's CANN-internal
implementation on the target Ascend NPU; $T_{\text{HW}}$, the analytical
Hardware-Anchored Performance (HAP) limit serving as a per-case lower bound;
and $T_{\text{cand}}$, the measured runtime of the candidate kernel. The
first two are computed once per case at benchmark-construction time;
$T_{\text{cand}}$ is collected at evaluation alongside the compilation and
accuracy flags.
Section~\ref{sec:precision-standard} specifies the
functional-correctness standard. Section~\ref{sec:hwlimit} motivates the
HAP design and defines $T_{\text{HW}}$.
Section~\ref{sec:scoring} gives the per-case performance score and assembles
the three axes into per-operator and suite scores.
Section~\ref{sec:reward-hacking} closes with reward-hacking surfaces and
the mitigation principles the evaluator deploys.

\subsection{Functional-Correctness Standard}
\label{sec:precision-standard}

CANN~Bench adopts a general and systematic precision standard as its functional-correctness gate.
The standard summarizes per-tensor agreement with the reference through two statistics on the elementwise relative error---its mean and its max.
If that first screen fails, the checker attributes the failing positions to the normal, small-value, or cancellation regions; only failures confined to the two numerically unstable regions are eligible for an ErrorCount-ratio fallback.
The standard also admits bounded per-operator relaxation when the underlying compute is unavoidably lossy.
The four engineering safeguards collected at the end of this subsection layer around the gate so that a kernel cannot pass it by exploiting the harness rather than by computing correctly.

\paragraph{Mean and max relative error.}
For each element $i$ at which both the kernel output $\hat y_i$ and the reference $y_i$ are finite, define the elementwise relative error (RE) as
\begin{equation}
e_i \;=\; \frac{|\hat y_i - y_i|}{|y_i| + 10^{-7}},
\end{equation}
where the $10^{-7}$ guard prevents a degenerate division when $y_i$ is exactly zero.
For dtype $d$ with default threshold $\tau_d$, the tensor clears the first accuracy screen when both the mean and the max of $\{e_i\}$ sit below their respective bounds:
\begin{equation}
\operatorname{mean}_i e_i \;<\; \tau_d \quad\text{and}\quad \max_i e_i \;<\; 10\,\tau_d.
\label{eq:re-gate}
\end{equation}
We refer to these two statistics as the mean RE and the max RE.
Positions containing $\pm\infty$ or $\mathrm{NaN}$ on either side are excluded from $\{e_i\}$ and handled separately by the special-value rule below.
When the first screen fails, any max-RE failure outside the small-value and cancellation regions is a normal-value failure; failures inside those two regions use the fallback rule below.
For multi-output operators every output tensor must clear the gate independently.

\paragraph{Per-dtype thresholds.}
Table~\ref{tab:precision-thresholds} lists the normal-range $\tau_d$ entries for the floating-point dtypes covered by the precision standard and mirrored in the evaluator.
The values are powers of two anchored to each format's mantissa width, so the bound tightens monotonically as the dtype gains precision; the max-RE bound is taken to be exactly $10\,\tau_d$ rather than the nearest power of two, matching the evaluator's implementation.
Integer dtypes are evaluated by exact equality by default, while an operator may declare a bounded integer tolerance for quantized outputs.
FP16, BF16, and FP32 are the active floating-point leaderboard gates in the current release; HiF32 and FP8 E4M3/E5M2 are implementation-backed threshold entries included for consistency with the region-specific tables in Appendix~\ref{app:precision-tables}. MXFP8, FP4, and MXFP4 remain roadmap formats (\S\ref{sec:status}).

\begin{table}[t]
\centering
\small
\renewcommand{\arraystretch}{1.20}
\begin{tabular}{@{}lcc@{}}
\toprule
Dtype & Mean-RE bound $\tau_d$ & Max-RE bound $10\,\tau_d$ \\
\midrule
FP16 & $2^{-10} \quad ( 9.77\times 10^{-4}$) & $ 9.77\times 10^{-3}$ \\
BF16 & $2^{-7 } \quad (7.81\times 10^{-3})$ & $ 7.81\times 10^{-2}$ \\
FP32 & $2^{-13} \quad  (1.22\times 10^{-4}$ )& $ 1.22\times 10^{-3}$ \\
HiF32 & $2^{-11} \quad( 4.88\times 10^{-4}$ )& $ 4.88\times 10^{-3}$ \\
FP8 E4M3 & $2^{-3  }  \quad ( 1.25\times 10^{-1})$ & $1.25$ \\
FP8 E5M2 & $2^{-2 }  \quad ( 2.50\times 10^{-1})$ & $2.50$ \\
\midrule
\multicolumn{3}{@{}l}{Integer dtypes: exact equality by default; bounded operator-declared tolerances are allowed for quantized outputs.} \\
\multicolumn{3}{@{}l}{Current leaderboard floating-point gates: FP16, BF16, and FP32; HiF32/FP8 rows are extension entries.} \\
\bottomrule
\end{tabular}
\caption{Per-dtype thresholds for the mean/max-RE gate. Each $\tau_d$ is a power of two anchored to the format's mantissa width; the max-RE bound is permitted to be ten times looser than the mean-RE bound so that an isolated outlier on a small subset of positions does not by itself fail an otherwise-correct tensor.}
\label{tab:precision-thresholds}
\end{table}

\paragraph{Three-region verdict.}
The mean/max-RE rule is the headline gate, but treats every elementwise position identically---which is the wrong default in two regions where relative error stops being a reliable signal on its own.
The evaluator therefore inspects the failing positions after the first screen and applies a region-specific fallback rather than rejecting the tensor outright.

\emph{Specific consideration on the small-value region.}
When the truncated reference $y_{\mathrm{trunc}}$ sits at or below the dtype's representable noise floor, the denominator $|y_i| + 10^{-7}$ in $e_i$ amplifies a single-ULP quantization difference into a spuriously large relative error even when the kernel is computing correctly.
This regime is routine in the tails of normalization layers, in softmax outputs that approach zero, and in gradient updates near a local minimum; flagging these positions as functional failures would penalize correct kernels for the dtype's own granularity rather than for any real error.

\emph{Specific consideration on the cancellation region.}
When two intermediate quantities of comparable magnitude subtract to produce a near-zero output, catastrophic cancellation (Kahan) wipes out the leading significant bits of the result.
A correct kernel can then emit $|\hat y_i| \approx 0$ whose elementwise relative error against the FP64 Golden is arbitrarily large---not because the arithmetic is wrong but because the result's dynamic range no longer matches the reference at the device dtype.
Attention score differences, mean-zero shifts in layer normalization, and residual deltas in optimizer state all routinely visit this regime.

For both regions the evaluator switches from relative error to an absolute-error count and compares the NPU count against a same-precision CPU run of the Golden:
\begin{equation}
\frac{\mathrm{ErrorCount}_{\mathrm{NPU}}}{\max\bigl(\mathrm{ErrorCount}_{\mathrm{CPU}},\,1\bigr)} \;\le\; 2,
\label{eq:ec-ratio}
\end{equation}
applied independently to the small-value and cancellation populations with the region-defining and counting thresholds tabulated in Appendix~\ref{app:precision-tables}.
The CPU same-precision reference is what makes this comparison meaningful: errors attributable to the dtype itself appear on both sides of the ratio and cancel out, so only kernel-induced contributions survive.
Once the first screen has failed, the tensor can still pass only if no failing position lies in the normal region and both special-region ratios clear their respective bounds.

\paragraph{Per-operator relaxation.}
The thresholds in Table~\ref{tab:precision-thresholds} are defaults rather than invariants.
The actual numerical loss of a correct implementation depends on the operator's compute graph: chained transcendental evaluations, repeated reductions, quantization, and stateful updates accumulate error in ways a single dtype-level bound cannot capture uniformly.
The benchmark therefore allows an operator's \texttt{proto.yaml} to override the effective threshold for any subset of its supported dtypes, decided case by case during operator authorship.
The published standard recommends capping floating-point relaxation at $10\times$ the default so that relaxations remain bounded; submissions are scored against the effective threshold declared by the operator, and that threshold is carried in the evaluation result so that a relaxed pass remains distinguishable from a strict one.

The four engineering safeguards below operate orthogonally to the threshold rule and apply to every dtype.

\paragraph{High-precision reference.}
By default, the CANN~Bench Golden path runs on CPU after upcasting floating-point inputs to FP64, which keeps the reference well above the device's native precision and prevents overflow or underflow during the reference computation itself from corrupting the truth.
The evaluator also supports native-precision CPU and NPU Golden modes for specialized benchmark integrations, but the default CANN configuration uses the FP64 CPU strategy.
Before $e_i$ is computed, the Golden output is cast to the kernel output's dtype, so that the threshold $\tau_d$ is interpreted at the device dtype rather than at the reference dtype.
Integer and boolean inputs retain their original dtype throughout, which keeps the path compatible with operators that reject a Double substitute (e.g.\ \texttt{Gcd} on integer inputs, \texttt{CrossEntropyLoss} on \texttt{int64} targets).

\paragraph{Special-value handling.}
$\mathrm{NaN}$ positions must match between the kernel output and the Golden.
Matching $\pm\infty$ entries with the same sign are excluded from the mean-RE and max-RE statistics in Eq.~(\ref{eq:re-gate}), while an infinity sign mismatch is a hard failure.
One-sided $\pm\infty$ entries are treated as saturation candidates: the infinite side is replaced by the maximum finite value of the output dtype and then subjected to the ordinary error gate, so saturation is not a blanket pass.

\paragraph{Opt-in two-trial fresh-input verification.}
The evaluator ships an opt-in second-pass mode for the accuracy gate.
When the accuracy-retry mode is enabled and a case clears the gate on the first trial, the evaluator re-samples a fresh input batch through the same case generator, perturbs every floating-point tensor by $0.01$ to defeat seed reuse, and re-runs Golden and candidate execution; only cases that clear both trials count toward functional correctness.
A submission that caches the first-trial output by data pointer or short-circuits its kernel after the first execution fails the second trial with stale or garbage results.
The default leaderboard pipeline currently invokes the single-trial path; the second pass is retained as an explicit hardening option.

\paragraph{Strict tensor type-identity.}
The evaluator validates each returned tensor through a strict type-identity check against the concrete \texttt{torch.Tensor} class rather than a permissive \texttt{isinstance} test, which rejects \texttt{FakeTensor} and other lazy-evaluation wrappers that can satisfy an \texttt{isinstance} check without performing any real computation.

\paragraph{Roadmap toward a more scientific standard.}
The mean/max-RE rule with three-region verdicts is the current functional-correctness gate; two follow-on directions are planned.
First, the benchmark will migrate from the ecosystem-operator standard now in use to Huawei's commercial operator precision standard as the latter stabilizes.
Second, the per-dtype tier will extend to cover the emerging low-precision formats on the precision roadmap in \S\ref{sec:status}.
Appendix~\ref{app:precision-tables} records the HiF32/FP8 starting thresholds already mirrored in the evaluator; MXFP8, FP4, and MXFP4 will require additional validation before they become leaderboard gates.
These refinements are co-developed with the CANN operator-base maintainers.

\subsection{Performance anchor: the Hardware-Anchored Performance (HAP) limit}
\label{sec:hwlimit}

The most natural design for a kernel-generation benchmark is to score a
submission by its speedup over a software reference. For a benchmark that
aims to remain stable across years, this design has a fundamental defect:
the reference drifts with framework and driver versions. A kernel that
scored $2{\times}$ over last quarter's CANN reference may score only
$1.4{\times}$ against a faster successor without any change to the kernel
itself. Longitudinal comparison loses meaning, and a published number
becomes a relative quantity that ages out within months.

Prior GPU benchmarks have also argued for hardware-derived anchors when
software-reference speedup is too volatile~\cite{lin2026solexecbench}. CANN~Bench
instantiates this idea through an Ascend-specific analytical anchor that we
refer to as the \emph{Hardware-Anchored Performance} (HAP) limit, denoted
$T_{\text{HW}}$. This HAP limit is the upper end of the score, while
$T_{\text{baseline}}$ remains the lower anchor so that progress is still framed
against the engineering reference users actually have access to. The
framework-dependent component appears only at the lower end of the score; the
upper end is anchored to the chip itself.

To make ``how fast can a kernel run'' precise on Ascend~910B2, two facts
must be acknowledged: 910B2 is a heterogeneous architecture in which six
execution units operate concurrently in time, and achievable performance
depends on a non-trivial space of implementation choices.

\paragraph{Six-unit decomposition.}
Let
\[
K \;=\; \{\,\text{cube},\ \text{vector},\ \text{mte1},\ \text{mte2},\ \text{mte3},\ \text{fixp}\,\}
\]
denote the set of execution units that operate concurrently within a single
AIC group. Cube performs matrix contractions in the
L0A$\times$L0B$\to$L0C accumulator; Vector handles elementwise, reduction,
and complex-activation work in UB; MTE2 and MTE3 are the GM$\leftrightarrow$UB
data-movement channels; MTE1 issues L1$\to$L0A/L0B prefetch for Cube; FixP
is the L0C$\to$GM write-back path that fuses simple epilogues such as bias,
scale, quant, clip, and ReLU. The 24 AIC groups on a 910B2 die replicate
this layout in parallel.

For a fixed kernel, let $t_k(\delta)$ denote the work time on unit $k$ under
implementation choice $\delta$ (work divided by peak rate), where $\delta$
ranges over active-core count, tile shape, schedule, buffer allocation, and
pipeline depth. Because the six units overlap in time, the device-side
dataflow time is bounded below by their maximum; the hardware limit is the
minimization of this maximum across the implementation space:
\begin{equation}
T_{\text{HW}} \;=\; \min_{\delta}\;\max_{k \in K}\; t_k(\delta).
\label{eq:hwlimit}
\end{equation}
$T_{\text{HW}}$ is the runtime a kernel would achieve under three
idealizations: every unit operating at peak rate, perfect overlap across all
six units, and zero host-side launch overhead. It depends only on the
operator's intrinsic FLOP and byte counts and on the published 910B2 peak
parameters; it remains constant across software-stack upgrades and CANN
releases. No real implementation can violate this bound.

\paragraph{Derivation procedure.}
$T_{\text{HW}}$ is computed once per case through an analytical procedure
grounded in 910B2 architectural documentation and published peak parameters.
The procedure takes as input the operator's mathematical definition, its
interface contract, and the case's shape and dtype. For each unit
$k \in K$ it estimates $t_k$ from the operator's FLOP and byte counts and
that unit's peak rate, takes the per-unit maximum, and minimizes over a
structured implementation-choice space $\delta$ that respects on-chip buffer
constraints (L1 capacity, L0A/L0B capacity, L0C accumulator size). The
procedure outputs $T_{\text{HW}}$ together with the dominant-bottleneck
label. The value of $T_{\text{HW}}$ is determined entirely by the hardware
specification: should the target hardware platform or its six-unit
characterization be revised, $T_{\text{HW}}$ is regenerated accordingly.

\paragraph{Per-case anchor pair.}
Each case is characterized by the analytical anchor pair
$(T_{\text{baseline}},\,T_{\text{HW}})$, which combines at evaluation time
with the candidate kernel's measured runtime $T_{\text{cand}}$. By the
physical meaning of the hardware limit, $T_{\text{HW}} \le T_{\text{baseline}}$
should hold on every case. Any observation that violates this inequality
triggers an evaluation review as an audit signal on either the analytical
lower-bound derivation or the timing measurement.

\paragraph{Bottleneck Attribution}
\label{paragraph:bottleneck}

\begin{figure}[t]
  \centering
  \begin{subfigure}[t]{0.32\textwidth}
    \centering
    \includegraphics[width=\textwidth]{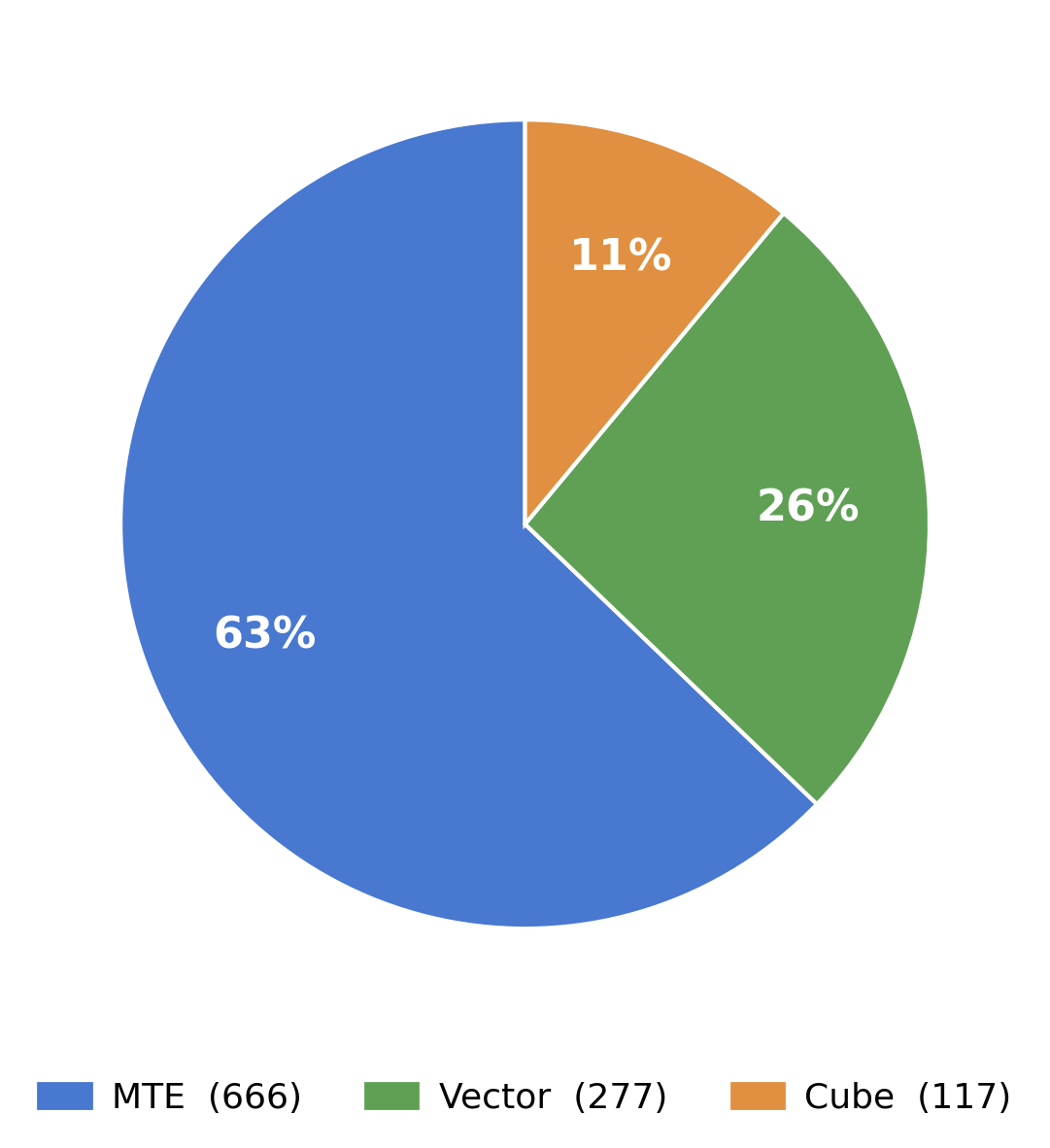}
    \caption{Dominant hardware bottleneck across the 1060 cases: MTE 63\%, Vector 26\%, Cube 11\%.}
    \label{fig:hw-bottleneck-pie}
  \end{subfigure}
  \hfill
  \begin{subfigure}[t]{0.64\textwidth}
    \centering
    \includegraphics[width=\textwidth]{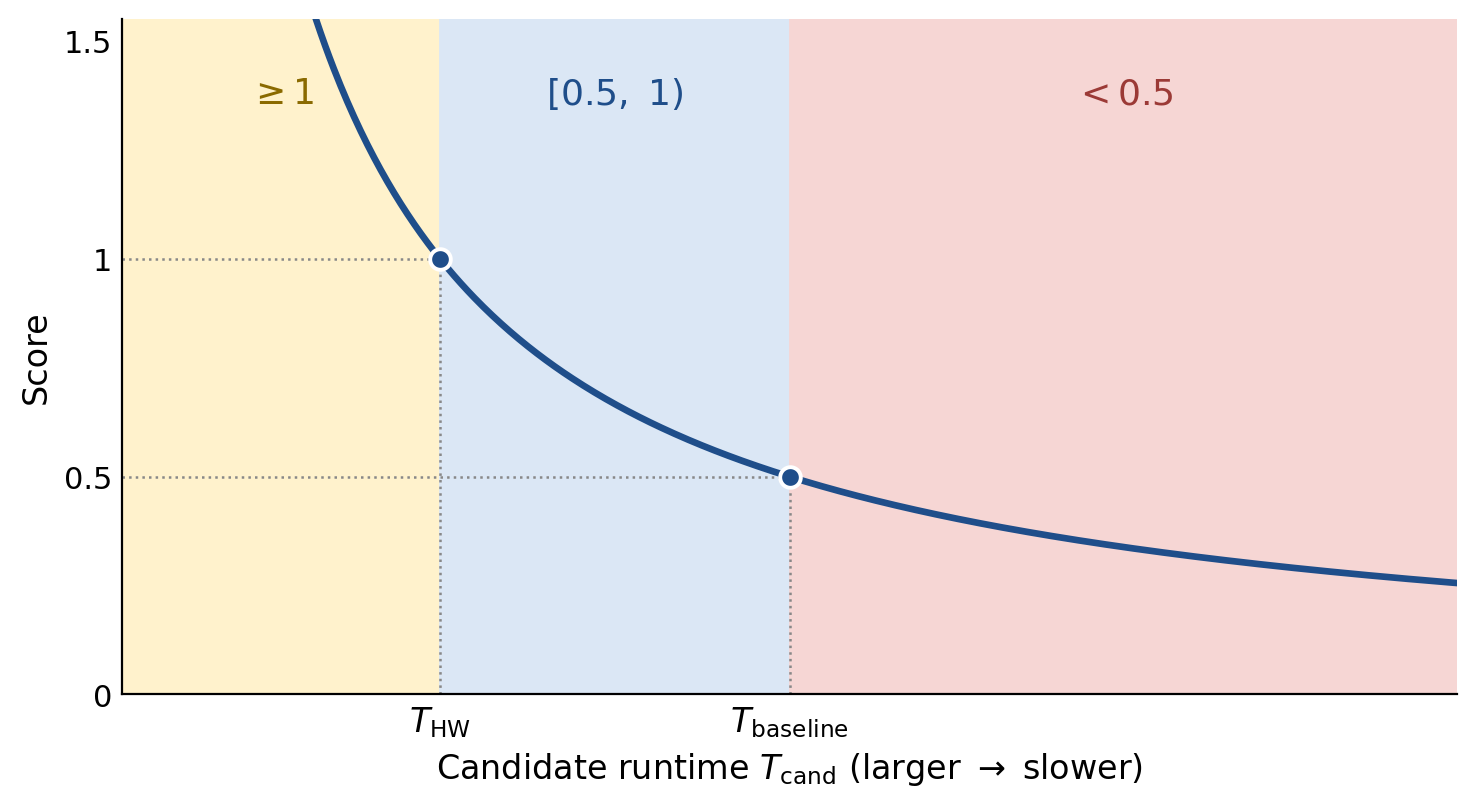}
    \caption{Per-case performance score versus candidate runtime $T_{\text{cand}}$: anchored at $1$ when $T_{\text{cand}}=T_{\text{HW}}$ and at $0.5$ when $T_{\text{cand}}=T_{\text{baseline}}$, decaying toward $0$ for slower kernels.}
    \label{fig:perf-score-curve}
  \end{subfigure}
\end{figure}

Each test case is annotated with the dominant hardware bottleneck that limits $T_{\text{HW}}$, classified into three categories: \emph{Vector} (Vector Core arithmetic), \emph{Cube} (Cube Core matrix computation), and \emph{MTE} (Memory Transfer Engine for data movement, rolling up MTE1/MTE2/MTE3 and the FixP writeback path).
As shown in Figure~\ref{fig:hw-bottleneck-pie}, MTE dominates at 63\% of the 1060 cases, with Vector at 26\% and Cube at 11\%. This distribution reflects the full Ascend pipeline workload, not just the compute engines.

\subsection{Per-case performance score and composite scoring}
\label{sec:scoring}
\paragraph{Per-case performance score}
Let $T_{\text{cand}}$ denote the candidate kernel's device-side measured
runtime. The per-case HAP score is defined as
\begin{equation}
\mathrm{Score}_i
\;=\;
\frac{T_{\text{baseline},i} - T_{\text{HW},i}}
{(T_{\text{cand},i} - T_{\text{HW},i}) + (T_{\text{baseline},i} - T_{\text{HW},i})}.
\label{eq:hw-score}
\end{equation}
The score has three anchor properties:
\begin{itemize}
\setlength\itemsep{2pt}
\item $T_{\text{cand}} = T_{\text{baseline}} \;\Rightarrow\; \mathrm{Score} = 0.5$
      (matching the reference implementation is the midpoint);
\item $T_{\text{cand}} = T_{\text{HW}} \;\Rightarrow\; \mathrm{Score} = 1$
      (matching the hardware limit is a perfect score);
\item $T_{\text{cand}} \to \infty \;\Rightarrow\; \mathrm{Score} \to 0$
      (slower kernels decay smoothly toward zero).
\end{itemize}
Figure~\ref{fig:perf-score-curve} plots $\mathrm{Score}_i$ as a function
of $T_{\text{cand}}$ for fixed $(T_{\text{baseline}}, T_{\text{HW}})$,
making the three anchor properties visible at a glance.
The midpoint design separates three regimes on a single bounded scale: worse
than the reference implementation ($\mathrm{Score} < 0.5$); better than
the reference implementation but sub-limit ($0.5 < \mathrm{Score} < 1$);
and at or beyond the hardware limit ($\mathrm{Score} \ge 1$). Submissions
that exceed the limit are not capped at $1$; the formula remains
well-defined and grows past $1$, but the case is flagged for review, since
$T_{\text{cand}} < T_{\text{HW}}$ implies one of three things: an
under-tight $T_{\text{HW}}$ derivation, a measurement-side
exploit~(\S\ref{sec:reward-hacking}), or a genuine modelling gap in the
analytical procedure. Each merits case-by-case investigation rather than
silent absorption into the score.

Compared with a raw baseline ratio
$T_{\text{baseline}}/T_{\text{cand}}$, equation~(\ref{eq:hw-score}) is
bounded, comparable across operators, and remains well-defined when
$T_{\text{baseline}}$ is already close to the hardware limit. The formula
awards similar credit for similar fractional progress within the
reference-to-limit window, regardless of how generous the reference happens
to be on the case at hand.

  \paragraph{Composite Score}                                                   
  \label{sec:composite-score}

  Equation~\eqref{eq:hw-score} grades a single test case along the performance 
  axis. A submission, however, is ranked at the operator level, and a
  leaderboard entry covers all evaluated operators. We therefore lift the       
  per-case score to a per-operator composite that also folds in compilation and
  functional correctness.

  Three signals enter the composite. At the operator level, a single compilation
   flag $\delta_{\mathrm{pass}} \in \{0, 1\}$ records whether the submitted
  source compiles; under the official ``one submission per operator'' protocol  
  this flag applies to the entire operator rather than to individual cases. At
  the case level, each case carries a functional indicator
  $\delta_{\mathrm{accuracy},i} \in \{0, 1\}$ (does the kernel match the golden
  reference under the per-dtype precision standard) and a per-case performance
  score $\mathrm{score}_i$ from Equation~\eqref{eq:hw-score}. The three axes
  are combined under linear weights $w_c = 0.2$, $w_f = 0.3$, $w_p = 0.5$ for
  compilation, functional correctness, and performance, with
  $\delta_{\mathrm{accuracy},i} \equiv 0$ whenever $\delta_{\mathrm{pass}} = 0$
  (a kernel that fails to compile cannot pass any case). The per-operator
  composite averages the weighted contributions across the operator's cases and
  is rescaled to a $0$--$100$ range:
  \begin{equation}
  \mathrm{EachOperatorScore}
  = \left[w_c \cdot \delta_{\mathrm{pass}} + \sum_{i \in \mathrm{cases}}
  \frac{\delta_{\mathrm{accuracy},i} \,(w_f + w_p \cdot \mathrm{score}_i)}{\mathrm{len}(\mathrm{cases})}\right]\cdot 100,
  \label{eq:composite}
  \end{equation}                                                                
  where the operator-level flag $\delta_{\mathrm{pass}}$ is replicated across
  every case in the sum, so the compilation term contributes a constant $w_c    
  \cdot \delta_{\mathrm{pass}}$ to the average regardless of how many cases the
  operator carries. The structure is therefore additive rather than             
  multiplicative: compilation contributes a single operator-wide $w_c$ floor on
  success, and each functionally passing case independently contributes a fixed
  $w_f$ floor plus a performance term $w_p \cdot \mathrm{score}_i$ that scales
  with the HAP-anchored fractional score of Equation~\eqref{eq:hw-score},
  before the whole quantity is normalised by $\mathrm{len}(\mathrm{cases})$ and
  lifted onto a $0$--$100$ scale. Per-level and benchmark-level totals are
  simple sums over the operators they cover:
  \begin{equation}
  \mathrm{Score}_{\text{Level-}N} = \sum_{\text{op} \in \text{Level-}N}         
  \mathrm{EachOperatorScore}_{\text{op}},                                       
  \qquad                                                                        
  \mathrm{Score}_{\text{benchmark}} = \sum_{N=1}^{4}                            
  \mathrm{Score}_{\text{Level-}N}.                                              
  \end{equation}
                                                                                
  This formulation keeps the three axes co-active. A submission that fails to   
  compile contributes nothing on any axis, since both the compilation term and
  every case-level term collapse to zero. A submission that compiles but fails  
  the functional gate on a given case forfeits both the $0.3$ floor and the $0.5
   \cdot \mathrm{score}_i$ term on that case while preserving the $0.2$
  compilation contribution and the contributions from any cases that do pass. A
  functionally correct but slow submission keeps the $0.3$ floor on every
  passing case and only loses (part of) the $0.5 \cdot \mathrm{score}_i$
  component. The weights $(w_c, w_f, w_p) = (0.2, 0.3, 0.5)$ are normalised to
  sum to $1$, so a fully compiled, functionally correct, hardware-limit-matching operator
  scores exactly $100$. The per-case ratio $w_p/w_f = 5/3$ keeps two extreme
  strategies on comparable footing: a fully correct submission at the baseline
  anchor ($\mathrm{score}_i = 0.5$ on every case) scores $75$ on the operator,
  while a partial-but-fully-optimised submission only overtakes that by clearing
   more than $11/16$ of its cases (roughly $14$ of the current $20$).

\subsection{Reward Hacking and Mitigation}
\label{sec:reward-hacking}

The composite score in Eq.~\eqref{eq:composite} is only meaningful if the measured outputs and runtimes come from the submitted NPU kernel. Kernel-generation benchmarks are vulnerable to evaluator loopholes: an optimizer can increase its score by routing computation to existing library kernels, hiding work outside the profiled window, caching outputs, or tampering with timing APIs rather than improving the generated kernel itself~\cite{lange2025robust,lin2026solexecbench}. CANN~Bench therefore treats anti-reward-hacking as part of the benchmark specification, not merely as post-hoc leaderboard review.

The Ascend evaluation stack exposes timing, synchronization, profiling, and custom-kernel entry points through the \texttt{torch.npu}/\texttt{torch\_npu} runtime and compiled Ascend C extensions~\cite{huawei_npu_cuda_alignment,swift_ascend_profiler,ascend_pytorch_repo}. CANN~Bench defends this surface through four principles: the submitted kernel must be the implementation being measured; candidate execution must stay on the target device; outputs must be materialized and validated case by case; and the measurement path must remain under evaluator control. Table~\ref{tab:reward-hacking-defenses} summarizes the deployed defense principles without exposing evaluator-internal trigger details.

\begin{table}[H]
\centering
\footnotesize
\setlength{\tabcolsep}{3.5pt}
\renewcommand{\arraystretch}{1.12}
\begin{tabularx}{\textwidth}{@{}>{\raggedright\arraybackslash}p{3.0cm} >{\raggedright\arraybackslash}X >{\raggedright\arraybackslash}p{5.6cm}@{}}
\toprule
\textbf{Surface} & \textbf{Exploit pattern} & \textbf{Mitigation principle} \\
\midrule
Framework or library delegation
& Route target semantics to existing framework, operator-library, or same-op CANN implementations instead of the submitted kernel.
& The submission contract forbids semantic delegation. Runtime checks enforce that measured work is attributable to the candidate kernel while preserving legitimate low-level primitives used inside that kernel. \\
\addlinespace[2pt]
CPU fallback or device egress
& Move tensors off the NPU, compute on the host, or return results whose performance cannot be attributed to NPU execution.
& Candidate execution is required to remain device-resident, and performance credit requires observable NPU-kernel execution for the submitted implementation. \\
\addlinespace[2pt]
State caching or hard-coded outputs
& Reuse stale outputs, memorize disclosed cases, or key behavior to input identity rather than implement the operator.
& Public/hidden case separation, per-case validation, subprocess isolation, and optional input refresh/address-diversity controls make score gains depend on general operator behavior. \\
\addlinespace[2pt]
Lazy or fake outputs
& Return wrapper objects that defer computation until validation or impersonate tensors.
& Output validation requires materialized tensors with the expected structure before timing and correctness results are accepted. \\
\addlinespace[2pt]
Timing and profiler tampering
& Modify synchronization, timing, or profiler behavior so the measured interval no longer reflects candidate execution.
& The evaluator protects the measurement path around submission installation and execution, rejecting runs whose timing or profiling surface is no longer trusted. \\
\bottomrule
\end{tabularx}
\caption{Reward-hacking surfaces covered by deployed CANN~Bench defenses. The table describes mitigation principles rather than evaluator trigger details.}
\label{tab:reward-hacking-defenses}
\end{table}

The same-op routing surface deserves separate treatment because it crosses the boundary between the submitted package and the evaluator's CANN installation. On dedicated evaluator machines, CANN~Bench can harden the environment so that leaderboard runs do not silently dispatch target semantics to preinstalled same-op binaries. This is an evaluator-machine policy, not a restriction on legitimate Ascend C intrinsics or other low-level building blocks compiled into the candidate kernel.

The remaining limitations are handled conservatively. Multi-stream submissions are restricted rather than fully ranked until stream visibility is stable across the supported runtime stack, and opaque precompiled binary loading remains subject to policy and manual review. Input refresh and address-diversity controls are useful hardening knobs when enabled by the evaluation profile, but the core security claim is simpler: performance credit is awarded only when the evaluator can validate the output and attribute the measured runtime to candidate NPU-kernel execution.

\section{Conclusion}
\label{sec:conclusion}

We introduced \textbf{CANN~Bench}, an Ascend-oriented operator-generation benchmark designed to give AI-generated kernels under Huawei's CANN stack a shared, vendor-aligned evaluation scale. The initial release ships 53 operators and 1060 test cases distributed across four difficulty levels (L1--L4), each operator accompanied by a self-contained reproducibility kit---\texttt{desc.md}, \texttt{proto.yaml}, \texttt{cases.csv}, and \texttt{golden.py}---and scored on the three axes that matter for kernel work on Ascend 910B2: compilation, functional correctness, and performance. Performance is graded between a PyTorch-on-Ascend baseline and an analytically derived per-case HAP limit, so absolute scores carry a hardware-grounded meaning rather than only a relative ranking. The harness, baseline-collection pipeline, and operator suite are co-located in the official CANN repository and released under the CANN Open Software License v2.0, with an online leaderboard prepared as the durable maintenance surface.

Relative to existing kernel-generation benchmarks, CANN~Bench occupies a position that none of the closest prior efforts simultaneously cover. Cross-platform suites such as KernelBench, TritonBench, and SOL-ExecBench target the NVIDIA GPU architecture and do not transfer mechanically to Ascend's tiling semantics and multi-level buffer structure; MultiKernelBench spans three hardware backends but treats Ascend as one slice with limited per-operator depth; and NPUKernelBench is co-evolved with a specific training recipe and includes a static-shape track where a single canonical input configuration governs an entire task. CANN~Bench differs in three concrete ways: golden references are vendor-authoritative and maintained alongside official CANN operator contracts; every operator carries a 20-case public split for local iteration plus an 80-case hidden split for leaderboard evaluation, which prevents fitting to disclosed inputs; and the harness combines specification-level invalid-behavior rules, runtime enforcement, measurement-integrity checks, and NPU-execution attribution so that performance credit remains tied to the submitted kernel.

We are explicit about the current limits of this release. Precision coverage is restricted to FP16, BF16, FP32, and INT8; FP8, HiF8, MXFP8, FP4, and MXFP4 are deliberate near-term extensions rather than shipped capabilities. Several reward-hacking surfaces remain intentionally conservative: multi-stream execution is restricted rather than fully ranked, because stream-level visibility across \texttt{torch\_npu} versions is still fragile; evaluator-machine hardening is a controlled leaderboard policy rather than an automatic property of every local install; and opaque pre-compiled binary loading is presently addressed by policy and manual review rather than automated provenance checks. Input refresh and address-diversity controls should be treated as run-profile-dependent hardening knobs unless enabled by the official leaderboard configuration. The benchmark also begins on a single hardware target (Ascend 910B2) and a single kernel programming surface, so coverage of newer Ascend silicon and additional DSLs is a near-term task rather than an established result.

Looking forward, CANN~Bench is intended to evolve with the Ascend ecosystem rather than freeze at release. Three extensions are scheduled: the online leaderboard portal will index and date every submission so that results stay directly citable; kernel-DSL coverage will extend beyond the initial Ascend-optimized path to Triton, PyPTO, TileLang, and other CANN-compatible paradigms under the same specification and scoring protocol; and precision coverage will expand to FP8, MXFP8, FP4, MXFP4, HiF8, and related low-precision formats as operators and baselines are validated. Versioned performance baselines, refreshable through a baseline-update utility planned for an upcoming release, ensure that reported speedups continue to track the current attainable reference rather than a stale snapshot as CANN and Ascend hardware progress. The operator set, golden implementations, harness, and documentation are open-source, and the project actively accepts contributions of new operators, test cases, baseline updates, and sub-leaderboards. By coupling vendor-authoritative golden references, the HAP anchor, a hidden-split evaluation protocol, and a deployed anti-reward-hacking harness inside the official CANN repository, CANN~Bench aims to serve as the shared, sustained evaluation scale that AI-generated kernels on Ascend have so far lacked, and to grow alongside the workloads, precisions, and silicon it is built to measure.

\appendix

\section{Region-specific precision thresholds}
\label{app:precision-tables}

This appendix collects the per-dtype thresholds used by the small-value and cancellation gates referenced in \S\ref{sec:precision-standard} (Eq.~\ref{eq:ec-ratio}). The values are taken from the ecosystem-operator precision standard and from the evaluator implementation; they are organized by region so that the main text can refer to a single ratio test rather than reproduce three threshold tables inline.

\paragraph{Small-value region.}
A position is classified as small-value when $|y_{\mathrm{trunc}}| < \tau^{\mathrm{sv}}_d$, where $y_{\mathrm{trunc}}$ is the FP64 Golden cast down to the device dtype and back to FP64 (i.e.\ the best value representable at the device precision). Within the region, a position is counted as an error when $|\hat y - y_{\mathrm{trunc}}|$ exceeds the per-dtype absolute-error tolerance $\varepsilon^{\mathrm{sv}}_d$. The NPU and CPU runs are both scored against the same $y_{\mathrm{trunc}}$, so that the ratio in Eq.~(\ref{eq:ec-ratio}) reflects only the kernel's contribution to the small-value error count.

\begin{table}[h]
\centering
\small
\renewcommand{\arraystretch}{1.20}
\begin{tabular}{@{}lcc@{}}
\toprule
Dtype & Small-value threshold $\tau^{\mathrm{sv}}_d$ & Absolute-error tolerance $\varepsilon^{\mathrm{sv}}_d$ \\
\midrule
FP16     & $2^{-11}$ & $2^{-16}$ \\
BF16     & $2^{-8}\phantom{0}$  & $2^{-16}$ \\
FP32     & $2^{-14}$ & $2^{-30}$ \\
HiF32    & $2^{-12}$ & $2^{-28}$ \\
FP8 E4M3 & $2^{-4}\phantom{0}$  & $2^{-6}\phantom{0}$  \\
FP8 E5M2 & $2^{-3}\phantom{0}$  & $2^{-5}\phantom{0}$  \\
\bottomrule
\end{tabular}
\caption{Small-value region thresholds. A position is in the small-value region when $|y_{\mathrm{trunc}}| < \tau^{\mathrm{sv}}_d$; within the region it contributes to $\mathrm{ErrorCount}$ when $|\hat y - y_{\mathrm{trunc}}| > \varepsilon^{\mathrm{sv}}_d$.}
\label{tab:small-value-thresholds}
\end{table}

\paragraph{Cancellation region.}
A position is classified as cancellation-affected when the kernel output is approximately zero, $|\hat y| < \tau^{\mathrm{cz}}_d$, while the Golden lies above the dtype's representable noise floor but inside the cancellation boundary, $\tau^{\mathrm{sv}}_d \le |y| < \tau^{\mathrm{cb}}_d$. The lower bound excludes positions already covered by the small-value rule. The cancellation count in Eq.~(\ref{eq:ec-ratio}) is the number of positions within the region whose relative error exceeds the max-RE bound $10\,\tau_d$ from Table~\ref{tab:precision-thresholds}; the NPU and CPU runs are evaluated against the same Golden so that the ratio isolates kernel-induced cancellation from dtype-floor cancellation.

\begin{table}[h]
\centering
\small
\renewcommand{\arraystretch}{1.20}
\begin{tabular}{@{}lcc@{}}
\toprule
Dtype & Cancel boundary $\tau^{\mathrm{cb}}_d$ & Cancel-zero threshold $\tau^{\mathrm{cz}}_d$ \\
\midrule
FP16     & $2^{-5}$ & $2^{-5}$ \\
BF16     & $2^{-3}$ & $2^{-3}$ \\
FP32     & $2^{-8}$ & $2^{-8}$ \\
HiF32    & $2^{-8}$ & $2^{-8}$ \\
FP8 E4M3 & $2^{-1}$ & $2^{-1}$ \\
FP8 E5M2 & $2^{0}\phantom{^{-1}}$  & $2^{0}\phantom{^{-1}}$  \\
\bottomrule
\end{tabular}
\caption{Cancellation region thresholds. A position is cancellation-affected when $|\hat y| < \tau^{\mathrm{cz}}_d$ and $\tau^{\mathrm{sv}}_d \le |y| < \tau^{\mathrm{cb}}_d$; the per-dtype values are anchored to the format's mantissa width so that the region captures losses of roughly one decade of significance around the dtype's precision floor.}
\label{tab:cancel-thresholds}
\end{table}

\bibliographystyle{unsrt}
\bibliography{references}

\end{document}